\documentclass{article}
\usepackage{ijcai18}

\usepackage{amssymb}
\usepackage{microtype}
\usepackage{graphicx}
\usepackage{ragged2e}
\usepackage{subfigure}
\usepackage{lipsum}
\usepackage{booktabs} 

\usepackage{times}
\usepackage{xcolor}
\usepackage{soul}
\usepackage[utf8]{inputenc}
\usepackage[small]{caption}

\usepackage{amsmath}
\usepackage{amssymb}
\usepackage{pifont}
\usepackage{multirow}
\usepackage{array, makecell}

\newcommand{\mypm}{\mathbin{\smash{%
\raisebox{0.35ex}{%
            $\underset{\raisebox{0.5ex}{$\smash -$}}{\smash+}$%
            }%
        }%
    }%
}

\title{Attentive Recurrent Tensor Model for Community Question Answering
}

\author{
Gaurav Bhatt$^1$, 
Shivam Sharma$^1$, 
Balasubramanian Raman$^1$, 
\\ 
$^1$ Indian Institute of Technology Roorkee \\
\{gauravbhatt.cs.iitr, shivamsharma.lmniit\}@gmail.com,
balarfma@iitr.ac.in
}

\begin{document}

\maketitle

\begin{abstract}
A major challenge to the problem of community question answering is the lexical and semantic gap between the sentence representations. 
Some solutions to minimize this gap includes the introduction of extra parameters to deep models or augmenting the external handcrafted features.
In this paper, we propose a novel attentive recurrent tensor network for solving the lexical and semantic gap in community question answering. We introduce token-level and phrase-level attention strategy that maps input sequences to the output using trainable parameters. 
Further, we use the tensor parameters to introduce a 3-way interaction between question, answer and external features in vector space. 
We introduce simplified tensor matrices with L2 regularization that results in smooth optimization during training.
The proposed model achieves state-of-the-art performance on the task of answer sentence selection (TrecQA and WikiQA datasets) while outperforming the current state-of-the-art on the tasks of best answer selection (Yahoo! L4) and answer triggering task (WikiQA).
\end{abstract}

\section{Introduction}
Deep learning models have been successfully used in community based question answering (CQA) problems such as answer sentence selection \cite{sigirChen2017benefit,pair_wise_level_rao2016noise,attn_tan}, best answer selection \cite{holo,yahoo1}, and answer triggering \cite{yang2015wikiqa}. 
A major challenge to these CQA tasks is the lexical and semantic gap for sentence matching. While most recent works explores the relationship between sentences in semantic space \cite{word_level,sent_level,sigirChen2017benefit}, some researchers have focused on interaction of QA pairs using an interactive layer \cite{ntn_cnn,holo,lex_decom}. The use of a bilinear layer such as a tensor helps to integrate the sentence modeling with the semantic modeling. The tensors are used to non-linearly compute the interaction between two vectors (or sentences) \cite{ntn}. The convolution neural tensor network (CNTN) uses the tensor on top of a convolution neural network (CNN) and has been shown to be effective for answer sentence selection in CQA \cite{ntn_cnn}. Some limitations to the use of tensor models might have to do with the less interaction of sentences in the semantic and lexical space. Additionally, the introduction of tensor matrices with deep learning models increases the trainable parameters. This leads to instability during the optimization of the learning model \cite{holo}.


In order to improve the sequence matching capabilities of deep learning models, some researchers have shown the effectiveness of attention mechanism \cite{attn_tan,yin2015abcnn,hermann2015teaching}. The attention alleviates the bottleneck of limited memory in deep models by introducing extra weight parameters that captures the relevance between two sequences. Other than attention, the use of feature engineering heuristics with deep learning architectures has proved to be effective for CQA \cite{sigirChen2017benefit,mohtarami2016sls}. Feature engineering introduces semantic and lexical relationships such as expressive rules 
which can support sentence matching to capture certain aspects of the information different than those captured by the underlying learning model \cite{mihaylov_nakov_2016semanticz,belinkov2015vectorslu,holo}.

In an attempt to match sentence pairs, some of the previous works follow point-wise ranking using a similarity matrix to learn lexical and semantic interaction among QA vectors \cite{severyn2015learning,lex_decom,holo}. These work concatenate the external features in the final merge layers. While these models compute interaction among QA vectors multiplicatively, they fail to express the interaction between QA vectors and external features.

In this paper, we solve the problem of the lexical and semantic gap along with the convergence problem faced by previous researchers.
We introduce a much interactive recurrent tensor model with attention (RTM) for sentence matching in CQA. 
We introduce two strategies for attention in RTM, depending upon the information content provided to the model. These are the token-level and phrase-level attention. 
Finally, we use three tensor matrices to compute a 3-way interaction between the question, answers, and external-features in the vector-space that further minimize the semantic and lexical gap. 
Instead of using multiple slices of tensors for each relation, we use a single tensor matrix with L2 regularization. 
With these simplified parameters, our model is able to overcome the convergence problems during training.

Finally, the main contributions of the paper can be summarized as 
\begin{enumerate} 
\item We present a novel recurrent tensor model (RTM) with phrase-level and token-level attention mechanism.
\item The RTM uses tensor matrices with L2 regularization for computing 3-way interaction between question, answers and external features for the task of CQA.
\item The presented idea is rather general to sentence matching and can be applied to a variety of NLP classification tasks.
\end{enumerate}

The rest of the paper is organized as follows. In section 2 we describe the problem statement in detail which is followed by the introduction of proposed techniques in section 3. We present experimental setup and results in section 4. This is followed by related work in section 5 and finally, our work is concluded in section 6.

\section{Community Question Answering}
CQA deals with problems where answers to textual queries are to be retrieved. Given a training corpus of relevant and irrelevant QA pairs, our task is to train a model that can learn to select the QA pairs according to their relevance. In this paper, we deal with 3 important tasks associated with CQA - answer sentence selection, best answer selection, and answer triggering. 

\subsection{Answer Sentence Selection}
In this task we are given a set of queries $q_i \in Q$ along with potential answers $a_i = \{a_{i_1},a_{i_2},...,a_{i_n}\}$. The answers comes with their relevancy judgments $\{y_{i_1},y_{i_2},...,y_{i_n}\}$, where the relevant answers are labeled as $1$ and the irrelevant answers are labeled as $0$. The goal of this task is to train a model that could generate an optimal order among the answers $a_i$ such that the relevant answers appear on the top.

\subsection{Best Answer Selection}
Similar to answer sentence selection, this task deals with queries along with some potential answers and the goal is to generate an optimal ordering. The difference lies in the relevancy judgment list - all the answers $a_i = \{a_{i_1},a_{i_2},...,a_{i_n}\}$ are relevant to the given query $q_i \in Q$. The elements of judgment list $\{y_{i_1},y_{i_2},...,y_{i_n}\}$ are labeled with a score indicating their relevance to the query. 

\subsection{Answer Triggering}
While answer sentence selection and best answer selection deals with the ranking of queries, the task of answer triggering is required for the (1) the detection of availability of at least one correct answer among the candidate sentences for every question; (2) if the first condition is satisfied, then one of the correct answer sentences from the candidate sentence set is selected.

\section{Recurrent Tensor Model}

\begin{figure*}[h!]
\centering\includegraphics[width=0.9\textwidth,height=0.4\textheight]{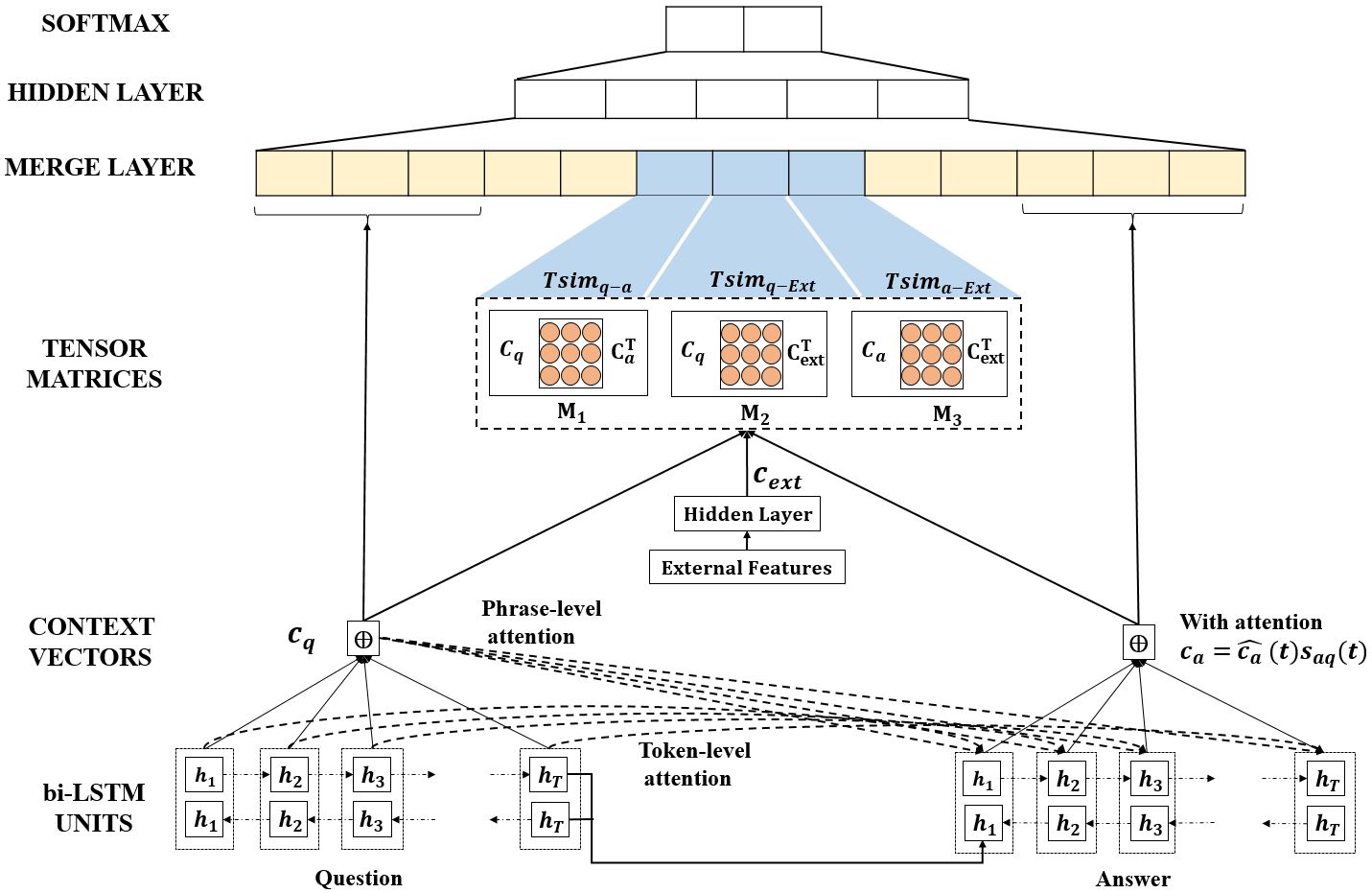}
\caption{The proposed recurrent tensor model with attention. The model uses either (a) - phrase level attention or (b) - token level attention, but not both.}
\label{fig:one}
\end{figure*}

\subsection{QA Encoding using attentive biLSTMs}
We use bi-direction Long Short Term Memories (biLSTMs) for generating vector encodings for textual QA pairs. 
Given a QA pair $(Q,A)$, where words in a question/answer is given by ${w}_{i} =  \{{w}_{1},{w}_{2},\cdots,{w}_{T}\}$, the biLSTM units are used to summarize the information from both directions of sequences.
The words $w_i$ are passed through an embedding matrix ${W}_{e}$ obtained from pre-trained word vector model giving vector representation ${v}_{ij} = {W}_{e}{w}_{ij}$. 
\begin{align}
{v}_{i}(t)&={W}_{e}{w}_{it}, t \in [1,T]; i \in [q,a] \label{eqn_9}
\end{align}
The biLSTM units contains the forward LSTM $\overrightarrow{f}$ that reads the question ${q}_{i}$ or answer ${a}_{i}$ from ${w}_{1}\ to\ {w}_{T}$ and a backward LSTM $\overleftarrow{f}$ that processes the same sequence from ${w}_{T}\ to\ {w}_{1}$. 
\begin{align}
\overrightarrow{{h}_{i}(t)}&=\overrightarrow{LSTM}({v}_{it}), t \in [1,T], i \in [q,a] \label{eqn_10}\\ 
\overleftarrow{{h}_{i}(t)}&=\overleftarrow{LSTM}({v}_{it}), t \in [1,T], i \in [q,a] \label{eqn_11}
\end{align}

The output of biLSTM units ($\widehat{c}_i$) after passing through pooling operation (shown as $\oplus$ in Figure 1) are called as the context vectors - $c_q$ for encoding of question and $c_a$ for the answer (see Figure 1). We use pooling operation to summarize the output of biLSTM units. The context vector $c_i$ for a given QA pair is obtained by concatenating the forward and backward LSTM outputs.

\begin{align}
\widehat{c}_{i}&=[\overrightarrow{{h}_{i}(t)} ; \overleftarrow{{h}_{i}(t)} \label{eqn_12}] \\
{c}_{i} &= \oplus (\widehat{c}_{i}), i \in [q,a]
\end{align}
where $\oplus$ is max or average pooling operation and the dimension of ${c}_i$ is $d$ for each QA pair. 

The answer context vector ($c_a$) is computed by adding attention weights to the answers, discussed in the following section.

\subsection{Attention}
A limitation to the use of biLSTMs is the long-range dependencies that arise due to longer sentences. To alleviate the problem of limited memory in biLSTM attention mechanism have been introduced \cite{attn_tan,hermann2015teaching}. The attention gives importance to certain words by opening a peeking window to question tokens while decoding the answers. Inspired by \cite{attn_tan,hermann2015teaching}, we introduce two types of attention mechanisms that can be used with RTM.

\subsubsection{Phrase-level attention}
In phrase-level attention, we use $c_q$ to compute the weighted sum of vectors, as shown in Figure 1. These weights define the degree to which the model will attend to question tokens when seen together in a summarized form. The value of $c_q$ is replicated across all time-stamps (T).
\begin{align}
{w}_{a,q}(t) &=W_{aw}\widehat{c}_a(t) + W_{qw}c_q
\end{align}

\subsubsection{Token-level attention}
In token-level attention, the model attends to each token of the question instead of seeing a summary. This means we use the biLSTM encoding for question prior to the pooling operation which is $\widehat{c}_{q}$ (see Figure 1). The value of $\widehat{c}_{q}$ is computed for all time-stamps (T). The weighted sum of vectors is computed as 
\begin{align}
{w}_{a,q}(t) &=W_{aw}\widehat{c}_a(t) + W_{qw}\widehat{c}_q(t)
\end{align}

Before passing the answer vector through pooling operation, the output is multiplied by a softmax weight, which is determined by the Eqn. (6) or (7) depending upon the attention mechanism used. The updated context vector $c_a$ for the answer tokens is given as:

\begin{align}
s_{a,q}(t) &= exp(\tilde{w}_{ws}tanh(w_{a,q}^T(t)))\\
c_a &= \oplus (\widehat{c}_a(t)s_{a,q}(t))
\end{align}
where $s_{a,q}(t)$ is the normalized attention token and $W_{aw}$, $W_{qw}$ and $\tilde{w}_{aw}$ are attention parameters.

\subsubsection{Interpretation of phrase-level and token-level attention}
The phrase-level RTM attends to the summary of a question which consists of only essential words. Whereas the token-level RTM give importance to each individual token.

Phrase-level attention has the advantage of skipping irrelevant words early, while there is a possibility that important words are skipped during the initial phase. Similarly, the token-level attention might have to deal with a lot of unimportant tokens but the chance of skipping important words is low.

Our work is different from \cite{attn_tan} as they focus entirely on phrase-level attention. The attention model introduced by \cite{hermann2015teaching} doesn't incorporate pooling which significantly improves the performance.

\subsection{External Features}
We use a total of 51 handcrafted features, some of which have not been used in previous studies. 

\textbf{Lexical and semantic features}.
We follow the work of \cite{yang2016beyond,yu2014deep_trec}, and compute features that cover topic relevance and semantic relatedness. These features include scores such as words overlap between QA pairs, length of QA pairs, BM25 score, weighted TF-IDF score, language model score, an exact match between QA pairs and DaleChall maximum similarity index. 

\textbf{Neural features}.
We create entity vectors for QA pairs using Word2vec, GloVe, and skip-thoughts, and use these vectors to calculate some set of features. These include distances and similarity measure such as \textit{Cosine distance, Manhattan distance, Jaccard similarity, Canberra distance, Euclidean distance, Minkowski distance} and \textit{Bray-Curtis distance}. 

\textbf{Readability}. These features lay emphasis to the text readability such as number of character per word (CPW) or number of syllable per word (SPW), number of words per sentence (WPS), number of complex words per sentence (CWPS), fraction of complex words (CWR), and Dale-Chall readability index \cite{yang2016beyond,sigirChen2017benefit}.


\subsubsection{Normalizing External Features}
To normalize the external features we pass them through a neural layer with $d$ neurons and apply a non-linearity $\alpha$. 
\begin{align}
c_{ext} = \alpha (w_h^T . Ext + b)
\end{align}
where $w_h\ \in\ 51*d$, $b\ \in\ 1*d$.

\subsection{Ranking using Tensor Matrices}
Our aim to use the tensor matrices is to find relationships $\mathbb{R}$ between three entities - ${c}_{q} \mathbb{R} {c}_{a}$, ${c}_{q}\mathbb{R} {c}_{ext}$ and ${c}_{a}\mathbb{R} {c}_{ext}$. To incorporate these three relations to our model we define tensor matrices $M$ indexed by $\mathbb{R} \in \{1,2,3\}$. RTM computes tensor similarity scores indicating how well the QA pairs are contextually related to each other: 
\begin{equation}
\begin{aligned}
\begin{matrix}Tsim_{q-a} \\ Tsim_{q-Ext} \\ Tsim_{a-Ext} \end{matrix}= f\Big( \begin{bmatrix}{{c}_{q}^T}{M}^{[1:k]}_{1}{c}_{a} \\
{{c}_{q}^T}{M}^{[1:k]}_{2}{c}_{ext} \\ {{c}_{a}^T}{M}^{[1:k]}_{3}{c}_{ext} \end{bmatrix}
\Big)
\end{aligned}
\end{equation}

where ${{M}_{R}}^{[1:k]} \in \mathbb{R}^{d*d*k}$ 
are the tensor matrices, $k$ is the slices of tensor matrices and $f$ is the non-linearity. The bilinear products 
are $k$ dimensional vectors where the entries are computed for every slice $i=1 \cdots k$. These bilinear products explores the relationship between the question, answer, and external-feature vectors. We observed that more slices of tensor matrices results in convergence problems. Our speculation is that each tensor matrix introduces extra tunable parameters, following which the model is unable to escape local solutions. For that reason we use only one tensor matrix for each relation as shown in Figure 1. We also show results varying the value of $k$ in the Experimental section.

\subsection{Merge Layer}
As shown in Figure 1, the tensor similarity scores are sandwiched between the context vectors.
\begin{align}
h_{merge} = [c_q;Tsim_{q-a};Tsim_{q-Ext};Tsim_{a-Ext};c_a]
\end{align}
In practice, combining context vectors for final score computation proves to be effective \cite{yin2015abcnn}. Each layer captures the properties of sequences at varying levels of abstraction which can be vital for a particular QA pair. 

The concatenated vector is passed through a hidden layer and a final softmax classification layer.

\subsection{Relation to tensor-based models}
The CNTN \cite{ntn_cnn} uses a neural tensor layer \cite{ntn} to computes a similarity score between a given QA pair: 

\begin{equation}
\begin{aligned} 
s({c}_{q},{c}_{a}) = {u}^{T}f\Big( {{c}_{q}^T}{M}^{[1:k]}{c}_{a}
 + {V} 
\begin{bmatrix} {c}_{q} \\ {c}_{a} \end{bmatrix} + b \Big)
\end{aligned}
\end{equation}
The CNTN model does not incorporate external features. Additionally, the $k$ slices of tensor matrices and linear weights $V$ increases the tunable parameters which results in convergence problem. 
The recently proposed LSTM with holographic projections \cite{holo} minimizes the number of tensor parameters while augmenting external features in the final layers but does not use attention.

Most of the tensor-based models fail to reduce the lexical and semantic gap due to lack of either attention mechanism or external features. The proposed model uses both of them which proves to be effective for CQA.

\subsection{Relation to point-wise ranking models}
The ranking model introduced by \cite{severyn2015learning,sigirChen2017benefit} use a point-wise method for ranking of short sentences. They used a similarity matrix with CNN to compute the similarity score:
\begin{align}
sim(c_q,c_a) = c_q^TMc_a
\end{align}
The external features are augmented in final layer. 
Similarly, the models given by \cite{attn_tan} uses cosine distance as point-wise similarity metric for QA context vectors.

In comparison, the computation of 3-way similarity score using regularized tensor matrices helps our model to compute the interaction between the questions, answers and their external features in a much interactive way.

\subsection{Training Objective}

We optimize the proposed model with cross-entropy loss for all three CQA tasks. The objective function for our model is given as 
\begin{eqnarray}  
J(\theta) = -\frac{1}{n} \sum_j \left[y_j \ln s_j + (1-y_j) \ln (1-s_j) \right] + \lambda{||\theta||}_{2}^{2}
\end{eqnarray}
where $\theta$ is the set of parameter for the proposed model and ${y}_i$ is the true score for a given QA pair. We use $L2$ regularization in hidden layers with $\lambda$ as the regularization constant. Moreover, we also use $L2$ regularization in all three tensor matrices separately.

\section{Experimentations}
In this section, we compare the performance of the proposed model with state-of-the-art techniques on CQA tasks. 

\begin{table}[b!]
\centering
\resizebox{5.8cm}{!}{%
\begin{tabular}{|l|l|}
\hline
\textbf{Hyperarameter} & \textbf{Value}\\
\hline
Embeddings          & Word2vec / GloVe\\
Embedding dimensions& 300 / 50\\
LSTM type           & Bidirectional      \\  
Pooling & Global Max\\
Merge layer neurons & 100                  \\ 
Optimizer           & Adam     \\
$\lambda$ - L2 regularization  & 0.000001     \\ 
Dropout       & 0.4      \\
Learning rate       & 0.001             \\ \hline
\end{tabular}
}
\caption{Hyper-parameters settings.}
\label{tab_hyperparameter}
\end{table}

\subsection{Dataset and Evaluation Metrics}
\textbf{WikiQA} \cite{yang2015wikiqa}: 
This dataset is used for answer sentence selection and answer triggering.
There are a total of  3047 questions with 2118 in the training set, 296 in the development set and 633 in the testing set. 
For both of these tasks, we truncate the entity tokens to 40 \cite{yang2015wikiqa}.

\begin{table*}
\centering
\begin{tabular}{c|cc|cc|ccc}
    \toprule
    \multirow{2}{*}{System} & \multicolumn{2}{|c|}{WikiQA} & \multicolumn{2}{|c}{TrecQA}&\multicolumn{3}{|c}{Yahoo! Manner L4}\\
    \cline{2-8}
     & MAP & MRR & MAP & MRR & MAP & MRR & P@1\\
    \midrule
CNN & 0.619 & 0.628 & 0.714 & 0.807 & 0.625 & 0.632 & 0.412\\
LSTM &0.624&0.614& 0.713 & 0.791 & 0.643 & 0.672 & 0.487\\
CNTN & 0.626 & 0.604 & 0.676 & 0.726 & 0.636 & 0.668 & 0.465 \\
    CNN+Cnt \cite{yang2015wikiqa} & 0.652 & 0.665 &-&-&-&-&-\\
    CNN+Feat \cite{severyn2015learning} &-&-&0.746 & 0.808&-&-&-\\
\midrule
	LSTM \cite{yahoo2} &-&-&-&-&-&  0.615 & 0.441\\
   biLSTM \cite{yahoo1} &-&-&-&-&-& 0.664 & 0.515\\
    Sent Level \cite{sent_level} & 0.693 & 0.709 & 0.762 & 0.830&-&-&-\\
    Word Level \cite{word_level} &  \textbf{0.709} & 0.723 & 0.755 & 0.825&-&-&-\\
   Pairwise \cite{pair_wise_level_rao2016noise} & 0.701 & 0.718 &  0.780 & 0.834 &-&-&-\\
   LSTM+Holograph \cite{holo} & - & - & 0.752 & 0.814 &-&0.734&0.556\\
  CNN+Ext \cite{sigirChen2017benefit} &0.700&0.714 & 0.782 & 0.837 &-&-&-\\
   ABCNN \cite{yin2015abcnn} & 0.692 & 0.710 & -&-& -&-&-\\
Lex decom. \cite{lex_decom} & 0.706 & 0.723 &-&-&-&-&-\\   
    \midrule
    RTM + phrase-level attention & 0.705 & 0.715 & \textbf{0.784} & 0.831 & 0.716 & \textbf{0.759}& \textbf{0.571}\\
     RTM + token-level attention & 0.701 & \textbf{0.726} & 0.774 & \textbf{0.839} & \textbf{0.721} & 0.741& 0.563\\
    \bottomrule
  \end{tabular}
\caption{Results on the tasks of answer sentence selection (WikiQA and TrecQA) and best answer selection (Yahoo! L4 Manner).}
\label{tab:correlation}
\end{table*}

\textbf{TrecQA}: 
The TrecQA dataset consists of QA pairs where answers to a given question are sorted according to their relevance. This dataset had been used widely for answer sentence selection task. The training set consists of a total of 5914 QA pairs having 89 unique questions, the development set with 1148 pairs having 65 unique questions and the test set consists of a total of 1442 pairs with 68 unique questions.

\textbf{Yahoo! webscope - L4}: 
Yahoo! manner\footnote{http://webscope.sandbox.yahoo.com/} dataset is publically available and contains $142,627$ questions from web forums. Each question is accompanied with a manually annotated best answer along with several relevant answers. For each question, we label the best answer as $1$ and other answers are labeled as $0$. We remove those questions that have multiple best answers and truncate all the answers to 300 tokens after preprocessing. We divide this dataset with 116031 questions in training set, 10000 in the development set and 15000 in the test set.

\textbf{Evaluation metrics}: For answer sentence selection we use Mean Average Precision (MAP) and Mean Reciprocal Recall (MRR) as the evaluation criteria. 
For best answer selection we use MAP and MRR along with Precision at 1 (P@1). 

MAP and MRR evaluate the relative ranks of correct answers in the candidate sentences of a question and hence are not suitable for evaluating the task of answer triggering. For this task, we use the precision, recall and F1 score. 

\subsection{Hyper-parameters and Baseline Methods}

The hyper-parameters used in our model is given in Table 2. 
As a common baseline across all the tasks we use basic CNN and LSTM. For CNN, we keep the values of hyper parameters as: number of filters = 100 (activation = relu), filter size = 50, global average pooling. For LSTM: number of hidden units = 100, dropout = 0.3, activation = relu.

We compare the proposed model with several state-of-the-art models - \cite{sigirChen2017benefit,holo,yin2015abcnn,word_level,sent_level,pair_wise_level_rao2016noise,miao2016neural,lex_decom,yahoo1}.

\subsection{Results and Discussion}
\subsubsection{CQA tasks}
The results of the answer sentence selection (WikiQA and TrecQA datasets) and best answer selection (Yahoo! L4) is shown in Table 2. The count features (CNN+Cnt) consists of weighted TF-IDF applied on words \cite{yang2015wikiqa} whereas the external features (CNN+Feat) used by \cite{severyn2015learning} consists of word overlap feature. The CNTN model is trained keeping the number of slices of the tensor as 4 (k=4). For our proposed model we report the results on two architectures - RTM with phrase-level attention and RTM with word-level attention. We experimented with two pre-trained word embeddings models - word2vec\footnote{300 dimensional embeddings - code.google.com/p/word2vec} and glove \footnote{50 dim. embeddings -  https://nlp.stanford.edu/projects/glove/}. We observed that with a smaller dimensional glove embeddings the model trains faster, while the performance is comparable to word2vec (only best results are reported).
CNN+Feat \cite{severyn2015learning} and lexical decomposition and composition (Lex decom) \cite{lex_decom} uses the point-wise similarity matrices for computing sentence similarity. ABCNN is the attention based CNN model \cite{yin2015abcnn}. 
SentLevel is the multiperspective model given by \cite{sent_level} and WordLevel denotes the model that computes pairwise word interaction \cite{word_level}. Pairwise is the triplet loss model that is trained on the sentence level and word level \cite{pair_wise_level_rao2016noise}. 

\begin{table}
\centering
\resizebox{\columnwidth}{!}{%
\begin{tabular}{|l|l|l|l|}
 \hline
 \textbf{Model} & {\textbf{Precision}} & {\textbf{Recall}} & {\textbf{F1}} \\
 \hline
 CNN &  23.89 & 33.21 & 27.78\\
 LSTM &  21.38 & 31.75 & 25.45\\
 CNTN &  24.14 & 37.80 & 29.46\\
 CNN+Cnt\cite{yang2015wikiqa} & 26.09 & 37.04 & 30.61\\
 CNN+All\cite{yang2015wikiqa} & 28.34 & 35.80 & 31.64 \\
 \hline
RTM+phrase-level attention & \textbf{29.32} & 43.93 & 35.16\\
RTM+token-level attention & 28.49 & \textbf{46.09} & \textbf{35.22}\\
 \hline
\end{tabular}
}
\caption{Results on the task of answer triggering (WikiQA dataset).}
\label{tab_res1}
\end{table}

\begin{figure*}
\centering\includegraphics[width=1\textwidth,height=0.18\textheight]{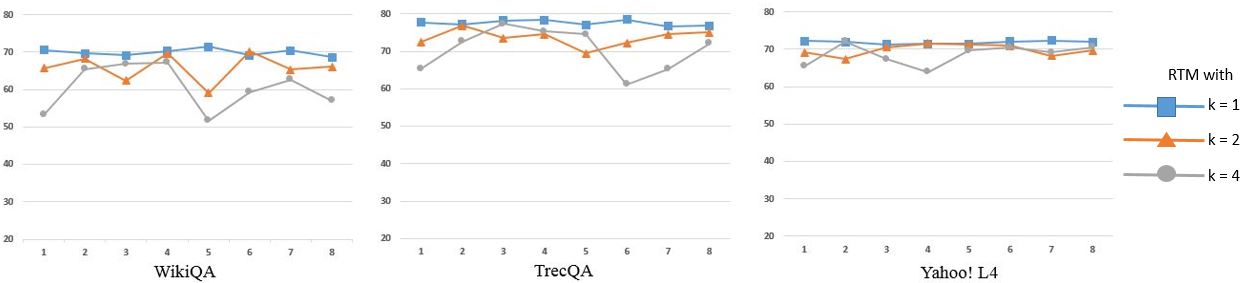}
\caption{Performance of RTM with varying tensor parameters (k = 1, 2 and 4). The vertical axis is the MAP scores on 10-fold cross-validation on TrecQA, WikiQA and Yahoo! L4 datasets.}
\label{fig:one}
\end{figure*}

RTM with attention achieves the state-of-the-art performance on answer sentence selection task. The performance of token-level attention is comparable to phrase-level attention on answer sentence selection and other tasks as well. The models that share similarity to the proposed model are CNN+Feat, LSTM+holograph, and CNTN. The RTM outperformed all these models by a clear margin. The CNTN model achieved a very low MAP and MRR scores on all three tasks. The reason being low semantic interaction among the QA pairs. The LSTM+holograph model \cite{holo} also use a simplified similarity bilinear matrix with circular correlation while the external features are augmented directly in the merge layer (this layer is similar to the merge layer from Figure 1). The introduction of token-level and phrase level attention along with 3-way simplified similarity matrices results in an improvement of MAP and MRR score when compared to these models. During experimentation, we observed that the global max pooling operation over the context vectors results in better performance. 

For best answer selection, our experimentation setup is similar to \cite{yahoo1,yahoo2} while slightly different from \cite{holo}. In their work, \cite{holo} label the best answer as 1, omitting the other given correct answers. They select 4 random answers and label them as 0. While we label the best answer as 1, the other correct answers are labeled 0. This makes our evaluation setup to be more challenging as for a given question the learning models have to differentiate between the best answer and contextually similar answers. Our proposed model outperformed all the state-of-the-art baselines and achieved a MAP score of 0.721, which is the highest score on this dataset.

For the task of answer triggering, we only use those questions that have at least one correct answer in its candidate list. 
The prediction of a question is done by considering the highest corresponding score given by the model. 
We find a threshold using the development data. If the score for a given answer is above this threshold and is labeled as correct, then this means we have a correct prediction.
In Table 2, the CNN+All is CNN model with count feature, the length of the question (QLen) and class of question as external features \cite{yang2015wikiqa}. The proposed model achieves a higher F1 score than the previous state-of-the-art baselines. We observe that this task is more challenging when compared to other CQA tasks as the highest F1 score across all models is relatively low.

\subsubsection{Simplified Tensor Parameters}
We performed experiments with varying the regularized tensor matrices in the RTM (for k = 1, 2 and 4). We use the 10-fold cross-validation setup for these experiments where each fold is computed from the training and development sets. For each run, we compute the MAP score as performance evaluation. The results on different parameter settings for the regularized tensor matrix is shown in Figure 2. With more tensor matrices (k$>$1) the MAP scores show a deviation range of $\approx \mypm 15 \%$ for all runs. The reason for this instability during optimization is the increase in trainable tensor parameters. The total number of tunable tensor matrices for RTM is $\mathbb{R}*k$ (where $\mathbb{R}$ = 3). For k=2, the total tensor matrices are 6, and so on. In order to keep a limit on excessive tunable parameters, we use simplified tensor matrices. As shown in Figure 2, the MAP scores of RTM with k=1 show the least range in deviation ($\approx \mypm 2 \%$) for all the runs.

\section{Related Work}

Commonly used ranking methods for CQA that have shown to achieve state-of-the-art performance are pairwise methods \cite{pair_wise_level_rao2016noise,word_level,sent_level} and point-wise methods \cite{severyn2015learning,lex_decom}. The use of tensor-based methods has also been used by previous researchers to achieve state-of-the-art performance on different CQA tasks \cite{ntn_cnn,holo}.

Several variants of deep learning models have been proposed by previous researchers to tackle the sentence matching problem in CQA \cite{attn_tan,lex_decom,yahoo1}. \cite{mohtarami2016sls} proposed a bag-of-vectors approach and used CNN and attention-based LSTMs to rank QA pairs. In other works, \cite{yin2015abcnn} used convolution neural network and \cite{miao2016neural} used recurrent neural networks with attention to find the relevance to answers for a given question.

Most of the previous work on automatic question answering for CQA sites show the advantage of feature engineering when used with deep models \cite{sigirChen2017benefit,mohtarami2016sls,yu2014deep_trec,severyn2015learning}. Feature engineering introduces additional semantic and lexical rules which help the underlying learning model to gain an overall increase in performance.
\cite{belinkov2015vectorslu,mihaylov_nakov_2016semanticz,tran2015jaist}.

\section{Conclusion}
In this paper, we presented a novel recurrent tensor model for sentence matching in CQA. The presented model leverages attention and 3-way similarity computation to minimize the lexical and semantic gap for CQA tasks.
Our experimental results demonstrate that the proposed RTM with attention outperformed most state-of-the-art baselines. We also observed that the performance of token-level attention is comparable to phrase-level attention. The lower relative F1 score in answer triggering task shows that there is room for improvement.

The presented idea in this paper is rather general to sentence matching and can be applied to a variety of tasks such as answer passage selection and knowledge-base QA.

\appendix

\bibliographystyle{named}
\bibliography{ijcai18}

\end{document}